%% file: main.tex
\documentclass{article}
\PassOptionsToPackage{numbers, compress}{natbib}
\usepackage[final]{neurips_2019}

\usepackage[utf8]{inputenc} % allow utf-8 input
\usepackage[T1]{fontenc}    % use 8-bit T1 fonts
\usepackage{hyperref}       % hyperlinks
\usepackage{url}            % simple URL typesetting
\usepackage{booktabs}       % professional-quality tables
\usepackage{amsfonts}       % blackboard math symbols
\usepackage{nicefrac}       % compact symbols for 1/2, etc.
\usepackage{microtype}      % microtypography
\usepackage{float}
\usepackage{graphicx}
\usepackage{ownstyles}
\usepackage{verbatim}
\usepackage{multirow}
\usepackage{tabu}
\usepackage{etoolbox}
\usepackage{wrapfig}
\usepackage{sidecap}
\renewcommand{\arraystretch}{1.5}

\graphicspath{{figures/}}

\newcommand{\capleft}{\textbf{(left)}\xspace}

\newcommand{\capright}{\textbf{(right)}\xspace}

\newcommand{\ltp}{\cite{frankle-2019-ICLR-the-lottery-ticket-hypothesis}\xspace}
\newcommand{\fcarb}{Frankle~\&~Carbin~\cite{frankle-2019-ICLR-the-lottery-ticket-hypothesis}\xspace}

\newcommand{\maskcrit}[1]{\ensuremath{\mathsf{#1}}\xspace}

\newcommand{\largefinal}{\maskcrit{large\_final}}
\newcommand{\smallfinal}{\maskcrit{small\_final}}
\newcommand{\largeinit}{\maskcrit{large\_init}}
\newcommand{\smallinit}{\maskcrit{small\_init}}
\newcommand{\lilf}{\maskcrit{large\_init\_large\_final}}
\newcommand{\sisf}{\maskcrit{small\_init\_small\_final}}
\newcommand{\maginc}{\maskcrit{magnitude\_increase}}
\newcommand{\movement}{\maskcrit{movement}}
\newcommand{\random}{\maskcrit{random}}
\newcommand{\largefinalss}{\maskcrit{large\_final\_same\_sign}}
\newcommand{\largefinalds}{\maskcrit{large\_final\_diff\_sign}}
\newcommand{\untrbase}{\maskcrit{untrained\_baseline}}
\newcommand{\trbase}{\maskcrit{trained\_baseline}}

\usepackage{algorithm,algcompatible,amsmath}
\algnewcommand\INPUT{\item[\textbf{Input:}]}%
\algnewcommand\OUTPUT{\item[\textbf{Output:}]}%

\algnewcommand{\IfThenElse}[3]{% \IfThenElse{<if>}{<then>}{<else>}
  \State \algorithmicif\ #1\ \algorithmicthen\ #2\ \algorithmicelse\ #3}
\usepackage{tikz}
\usetikzlibrary{decorations.pathreplacing,calc}

\newif\ifcomments

\commentsfalse
\ifcomments
\newcommand{\comments}[1]{#1}
\else
\newcommand{\comments}[1]{}
\fi
\usepackage{xr}
\makeatletter
\newcommand*{\addFileDependency}[1]{% argument=file name and extension
  \typeout{(#1)}
  \@addtofilelist{#1}
  \IfFileExists{#1}{}{\typeout{No file #1.}}
}
\makeatother

\newcommand{\titl}{Deconstructing Lottery Tickets:\\Zeros, Signs, and the Supermask}
\title{\titl}

\makeatletter
\let\@fnsymbol\@arabic
\makeatother
    
\input{authors}

\begin{document}

\maketitle

\begin{abstract}
  The recent ``Lottery Ticket Hypothesis'' paper by Frankle \& Carbin
  showed that a simple approach to creating sparse networks (keeping the
  large weights) results in models that are trainable from scratch,
  but only when starting from the same initial weights. The
  performance of these networks often exceeds the performance of the
  non-sparse base model, but for reasons that were not well
  understood. In this paper we study the three critical components of
  the Lottery Ticket (LT) algorithm, showing that each may be varied
  significantly without impacting the overall results. Ablating these
  factors leads to new insights for why LT networks perform as well as they do. We
  show why setting weights to zero is important, how signs are all
  you need to make the reinitialized network train, and why masking
  behaves like training. Finally, we discover the existence of
  Supermasks, masks that can be applied to an untrained, randomly
  initialized network to produce a model with performance far better
  than chance (86\% on MNIST, 41\% on CIFAR-10).

\end{abstract}

\section{Introduction}
\seclabel{introduction}

Many neural networks are over-parameterized
\citep{dauphin-2013-arXiv-big-neural-networks-waste,denil-2013-NIPS-predicting-parameters-in-deep},
enabling compression of each layer
\cite{denil-2013-NIPS-predicting-parameters-in-deep,wen-2016-arXiv-learning-structured-sparsity,han-2015-ICLR-deep-compression:-compressing}
or of the entire network
\cite{li-2018-ICLR-measuring-the-intrinsic-dimension}.
Some compression approaches enable more efficient computation by
pruning parameters, by factorizing matrices, or via other tricks
\cite{
  han-2015-ICLR-deep-compression:-compressing,
  NIPS1992_647,
  NIPS1989_250,
  li-2017-ICLR-pruning-filters-for-efficient,
  louizos-2017-arXiv-bayesian-compression-for-deep,
  luo-2017-ICCV-thinet-a-filter-level,
  molchanov-2017-ICLR-pruning-convolutional-neural,
  wen-2016-arXiv-learning-structured-sparsity,
  yang-2017-CVPR-designing-energy-efficient-convolutional,
  yang-2015-CVPR-deep-fried-convnets}.
Unfortunately, although sparse networks created via pruning often work well, training sparse networks directly often
fails, with the resulting networks underperforming their dense counterparts
\cite{li-2017-ICLR-pruning-filters-for-efficient,han-2015-ICLR-deep-compression:-compressing}.

A recent work by \fcarb was thus surprising to many researchers when it presented a simple algorithm
for finding sparse subnetworks within larger networks that \emph{are} trainable from scratch.
Their approach to finding these sparse, performant networks is as follows:
after training a network, set all weights smaller than some threshold to zero, pruning them (similarly to other pruning approaches \cite{NIPS2015_5784,han-2015-ICLR-deep-compression:-compressing,li2016pruning}), rewind the rest of the weights to their initial configuration, and then retrain the network from this starting configuration but with the zero weights frozen (not trained). Using this approach, they obtained two intriguing results.

First, they showed that the pruned networks performed well. Aggressively pruned networks (with 95 percent to 99.5 percent of weights pruned) showed no drop in performance compared to the much larger, unpruned network. Moreover, networks only moderately pruned (with 50 percent to 90 percent of weights pruned) often outperformed their unpruned counterparts.
Second, they showed that these pruned networks train well only if they are rewound to their initial state, including the specific initial weights that were used. Reinitializing the same network topology with new weights causes it to train poorly. As pointed out in \ltp, it appears that the specific combination of pruning mask and weights underlying the mask form a more efficient subnetwork found within the larger network, or, as named by the original study, a lucky winning ``Lottery Ticket,'' or LT.

% Second, compelling though these results were, the characteristics of the remaining network structure and weights were just as interesting. Normally, if you take a trained network, re-initialize it with random weights, and then re-train it, its performance will be about the same as before.
% %But with the skeletal Lottery Ticket (LT) networks, this does not hold.
% But with their pruned networks, they showed that this property does not hold.
% The network trains well only if it is rewound to its initial state, including the specific initial weights that were used. Reinitializing the same network topology with new weights causes it to train poorly. As pointed out in \ltp, it appears that the specific combination of pruning mask and weights underlying the mask form a lucky, trainable subnetwork found within the larger network, or, as named by the original study, a lucky winning ``Lottery Ticket,'' or LT.

%\removed{Their associated 
% ``Lottery Ticket Hypothesis'' states that
%inside the original, large network, a
%subnetwork together with its initialization forms the winning ticket that results in a more efficient network.}

While \fcarb clearly demonstrated LT networks to be effective, it raises many intriguing questions about the underlying mechanics of these subnetworks. What about LT networks causes them to show better performance? Why are the mask and the initial set of weights so tightly coupled, such that re-initializing the network makes it less trainable? Why does simply selecting large weights constitute an effective criterion for choosing a mask? 
We attempt to answer these questions by exploiting the essential steps in the lottery ticket algorithm, described below:

\begin{SCfigure}[20]  % keep this number large I think
\centering
\caption{Different mask criteria can be thought of as segmenting the 2D ($w_i =$ initial weight value, $w_f =$ final weight value) space into regions corresponding to mask values of $1$ vs $0$. The ellipse represents in cartoon form the area occupied by the positively correlated initial and final weights from a given layer. The mask criterion shown, identified by two horizontal lines that separate the whole region into mask-1 (blue) areas and mask-0 (grey) areas, corresponds to the \largefinal criterion used in \ltp: weights with large final magnitude are kept and weights with final values near zero are pruned. 
}
\label{fig:kd_large_final_detailed}
\includegraphics[width=0.26\textwidth]{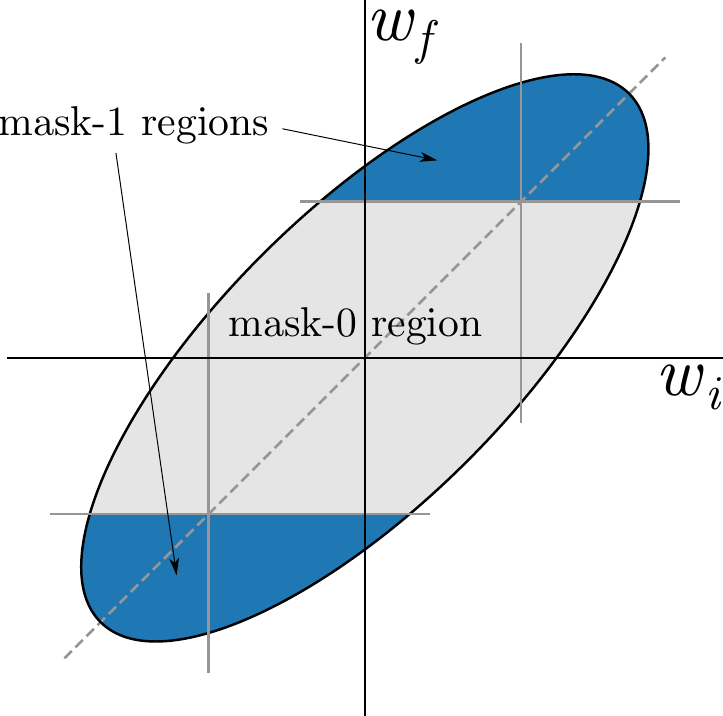} % change this to change width of image
\end{SCfigure}

%We begin by briefly describing the lottery ticket algorithm:

%more formally in Algorithm 1. (Note: copied in some cases verbatim from \ltp; for clarity we have changed as little as possible, though we present the masks in a slightly more general context to support the explorations done later.)

%Here
%$\vec{x}$ is a network's input vector,
%$\vec{w} \in \mathcal{R}^n$ is a vector containing all $n$ parameters in a network,
%$\vec{w}_i$ the initial (untrained) parameter vector,
%$\vec{w}_f$ the final (trained) parameter vector,
%$w_i$ and $w_f$ individual elements of those respective vectors,
%$\vec{m}$ a mask $\in \{0,1\}^n$.
%Let $M(w_i, w_f)$ be a \emph{mask criterion} function from $\mathcal{R}^2$ to $\mathcal{R}$ that maps (initial, final) weight pairs to a score, where weight pairs with higher scores are preferentially kept (masked to 1) over weights with lower scores that are masked to 0.
%%$m_i$ an element of that mask.
%Note: some abuse of notation: below, where appropriate, $w$ may apply to vector or individual. s\jl{why not use W for vector and w for individual?}

\begin{enumerate}
  \setcounter{enumi}{-1}
  \item Initialize a mask $m$ to all ones. Randomly initialize the parameters $w$ of a network \mbox{$f(x; w \odot m)$} %== f(x; m0 * w0).

  \item Train the parameters $w$ of the network $f(x; w \odot m)$ to completion. Denote the initial weights before training $w_i$ and the final weights after training $w_f$.

  \item \emph{Mask Criterion.} Use the mask criterion $M(w_i, w_f)$ to produce a masking score for each currently unmasked weight.
    %--- those corresponding to an associated value of 1 in the mask vector.
    Rank the weights in each layer by their scores, set the mask value for the top $p\%$ to 1, the bottom $(100-p)\%$ to 0, breaking ties randomly. Here $p$ may vary by layer, and we follow the ratios chosen in \ltp, summarized in     \tabref{arch}.
    %in pruning all layers but the last at an equal rate, with the last layer pruned at half the rate.
    In \ltp the mask selected weights with large final value
    %, here denoted \largefinal and
    corresponding to $M(w_i, w_f) = |w_f|$. 
    % In \secref{masks} we consider other mask criteria.
    
  \item \emph{Mask-1 Action.} Take some action with the weights with mask value 1. In \ltp these weights were reset to their initial values and marked for training in the next round. 
%   We consider this and other mask-1 actions in \secref{oneaction}.
    
  \item \emph{Mask-0 Action.} Take some action with the weights with mask value 0. In \ltp these weights were pruned: set to 0 and frozen during any subsequent training. 
%   We consider this and other mask-0 actions in \secref{zeroaction}.
    
  \item Repeat from 1 if performing iterative pruning.
\end{enumerate}

In this paper we perform ablation studies along the above three dimensions of variability, considering alternate mask criteria (\secref{masks}), alternate mask-1 actions (\secref{oneaction}), and alternate mask-0 actions (\secref{zeroaction}). These studies in aggregate reveal new insights for why lottery ticket networks work as they do.
Along the way we discover the existence of Supermasks---masks that 
produce above-chance performance when applied to untrained networks (\secref{supermask}). We make our code available at \url{https://github.com/uber-research/deconstructing-lottery-tickets}.
% , and directly training the Supermask offers an additional remarkable boost 

%In this paper, we argue that \emph{Mask Criterion}, \emph{Mask-1 Action} and \emph{Mask-0 Action} are the three key defining factors in almost any pruning algorithm that follows this general framework. Authors in \ltp explored some combinations of options along each of the three dimensions and obtained insightful results. The purpose of this paper is to perform much more extensive experiments and ablation studies in these three dimension, to shed some light in the importance, robustness, as well as implications of all the choices.
%In \ltp, authors explored choices of the mask criterion, mask-0 action, and mask-1 action that led to curiously good results. In the following sections, we consider alternate choices along each of these dimensions to expose which factors are important and which incidental to good performance.

\section{Mask criteria}
\seclabel{masks}

We begin our investigation with a study of different \emph{Mask Criteria}, or functions that decide which weights to keep vs. prune.
%How are masks produced? Generally, a mask is obtained from a function of the weight values.
In this paper, we define the mask for each individual weight as a function of the weight's  values both at initialization and after training: $M(w_i, w_f)$. We can visualize this function as a set of decision boundaries in a 2D space as shown in \figref{kd_large_final_detailed}.
In \ltp, the mask criterion simply keeps weights with large final magnitude;
we refer to this as the \largefinal mask,
$M(w_i, w_f) = |w_f|$.

%$M_\mathrm{large\_final}(w_i, w_f) = |w_f|$.
%As shown in \figref{kd_large_final_detailed}, the tips of the ellipse, which symbolize the entire region of network weights, are kept, and the remaining middle are pruned according to this criterion.

%Why does $M_\mathrm{large\_final}$ work so well? Are other mask strategies as effective? In this section, we explore a number of different mask criteria.

We experiment with mask criteria based on final weights (\largefinal and \smallfinal), initial weights (\largeinit and \smallinit), a combination of the two (\lilf and \sisf), and how much weights move (\maginc and \movement). We also include \random as a control case, which chooses masks randomly. These nine masks are depicted along with their associated equations in \figref{masks}. 
Note that the main difference between \maginc and \movement is that those weights that change sign are more likely to be kept in the \movement criterion than the \maginc criterion.

\newcolumntype{L}{>{\centering\arraybackslash}m{0.095\linewidth}}
\begin{figure}[t]
  \vskip 0.15in
\begin{center}
\begin{small}
{\setlength{\tabcolsep}{4pt}
\hspace*{-.75em}\begin{tabular}{LLLLLLLLL}
\toprule
large final &
small final &
large init &
small init &
large init large final &
small init small final &
magnitude increase &
movement &
random \\
%
%$M_{\mathrm{large\_final}}$ &
%$M_{\mathrm{small\_final}}$ &
%$M_{\mathrm{large\_init}}$ &
%$M_{\mathrm{small\_init}}$ &
%$M_{\mathrm{large\_init\_large\_final}}$ &
%$M_{\mathrm{small\_init\_small\_final}}$ &
%$M_{\mathrm{magnitude\_increase}}$ &
%$M_{\mathrm{movement}}$ \\
%
$|w_f|$ &
$-|w_f|$ &
$|w_i|$ &
$-|w_i|$ &
\scalebox{.65}{$min(\alpha|w_f|, |w_i|)$} &
\scalebox{.60}{$-max(\alpha|w_f|, |w_i|)$} &
\scalebox{.9}{$|w_f|-|w_i|$} &
$|w_f - w_i|$ &
0\\
\includegraphics[width=1.0\linewidth]{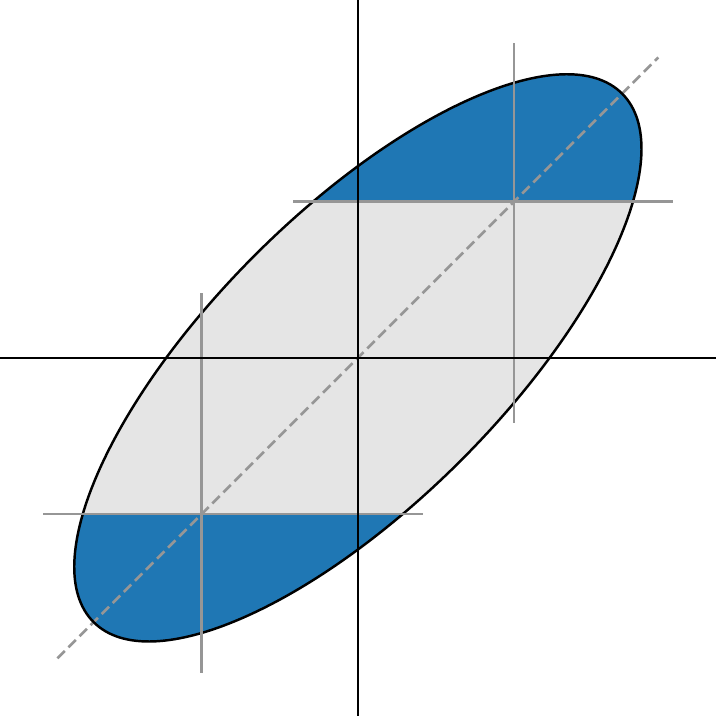} &
\includegraphics[width=1.0\linewidth]{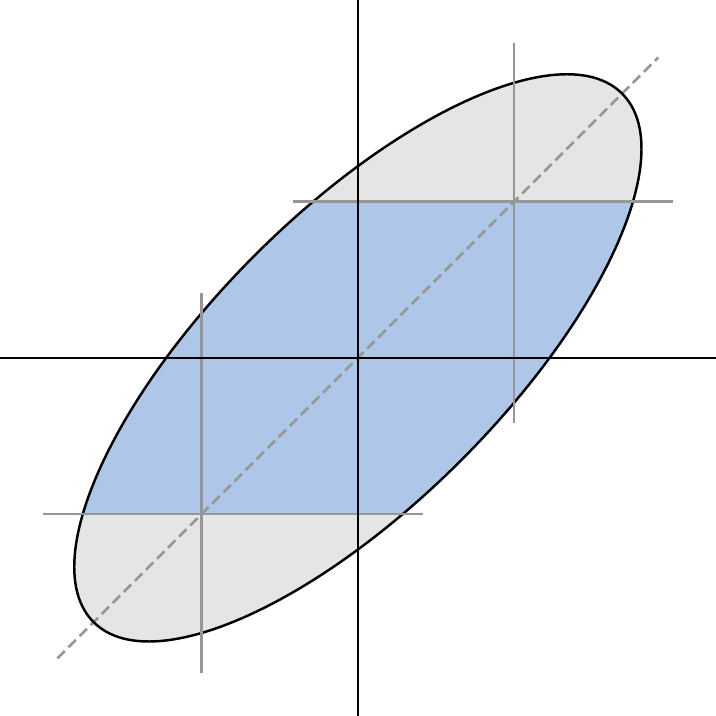} &
\includegraphics[width=1.0\linewidth]{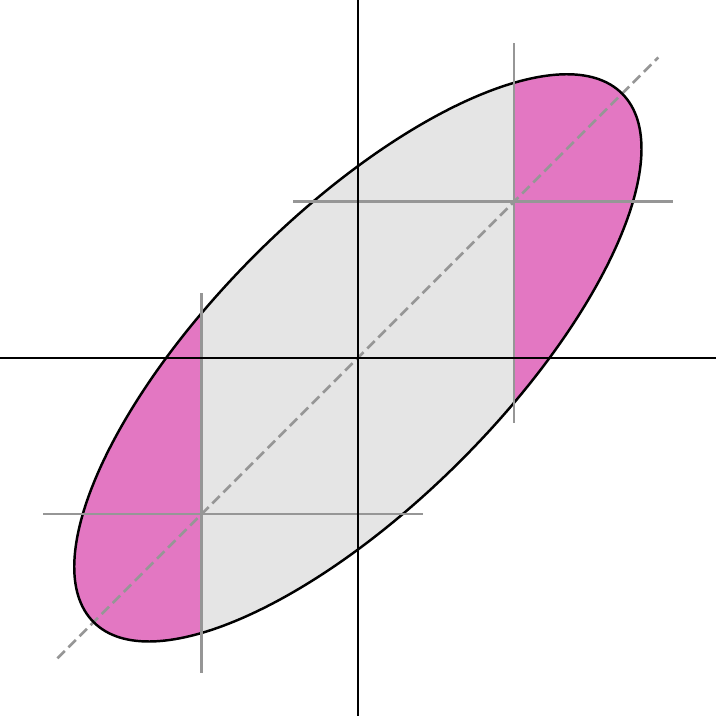} &
\includegraphics[width=1.0\linewidth]{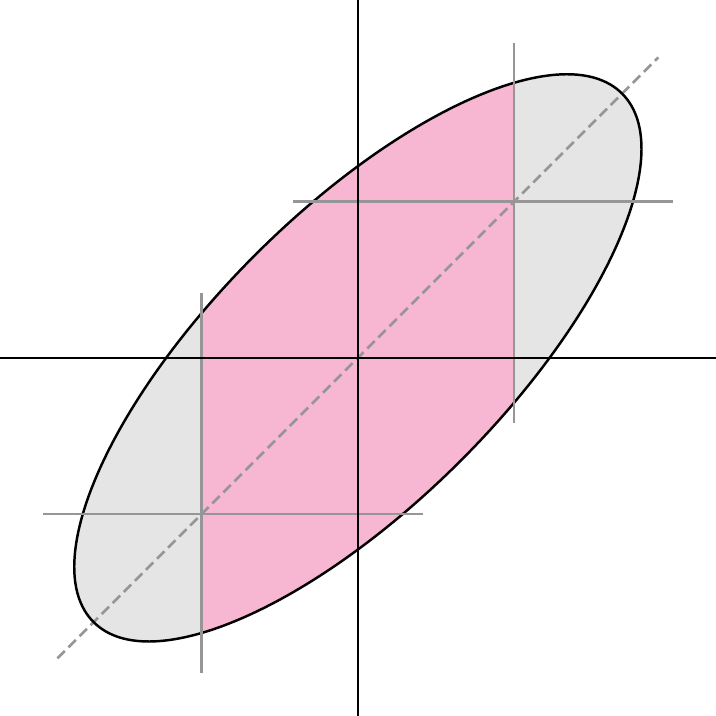} &
\includegraphics[width=1.0\linewidth]{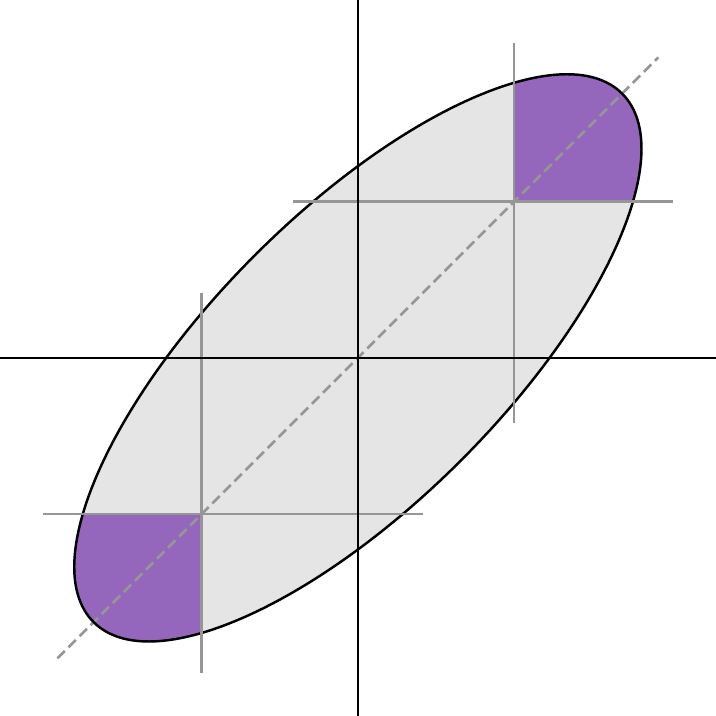} &
\includegraphics[width=1.0\linewidth]{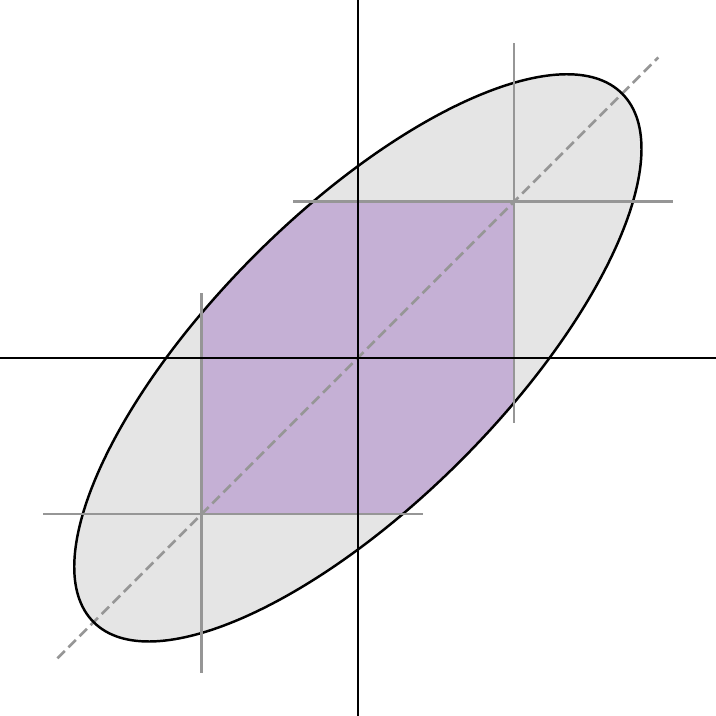} &
\includegraphics[width=1.0\linewidth]{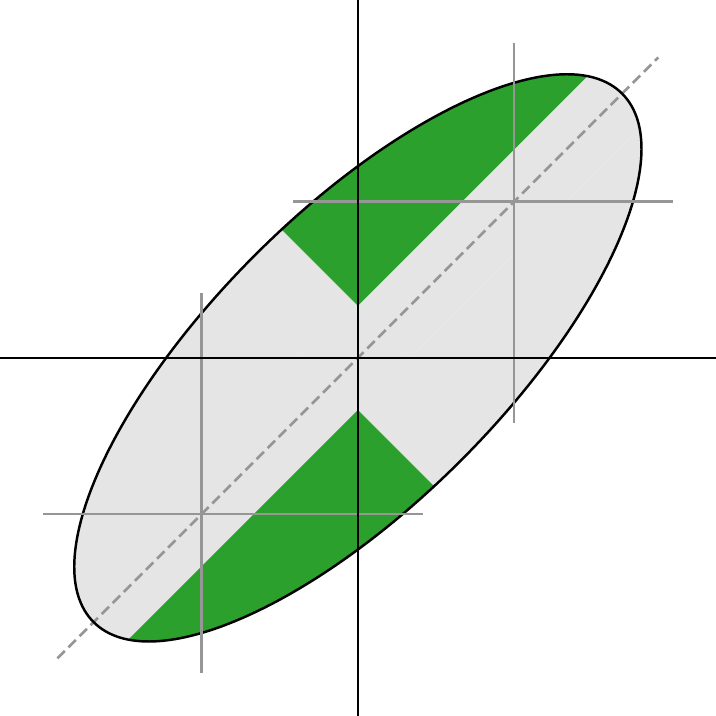} &
\includegraphics[width=1.0\linewidth]{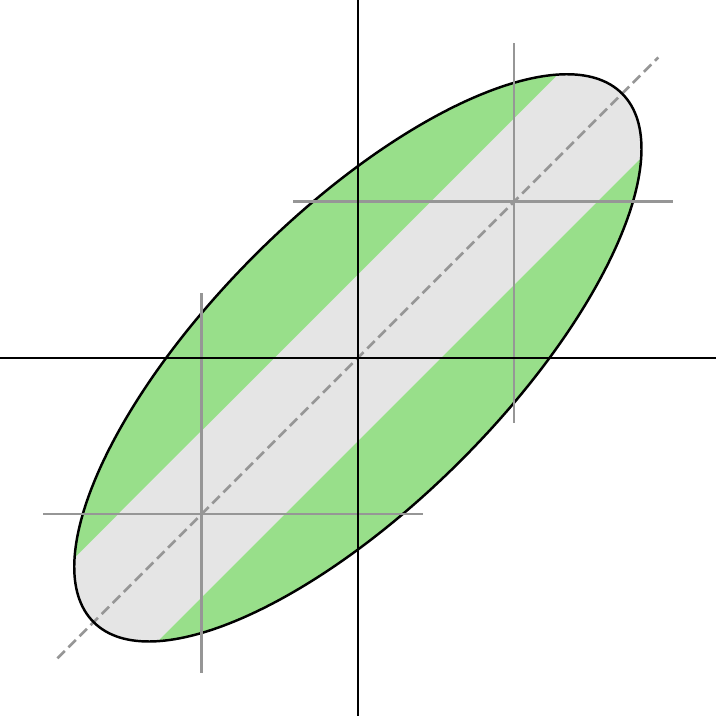} &
\includegraphics[width=1.0\linewidth]{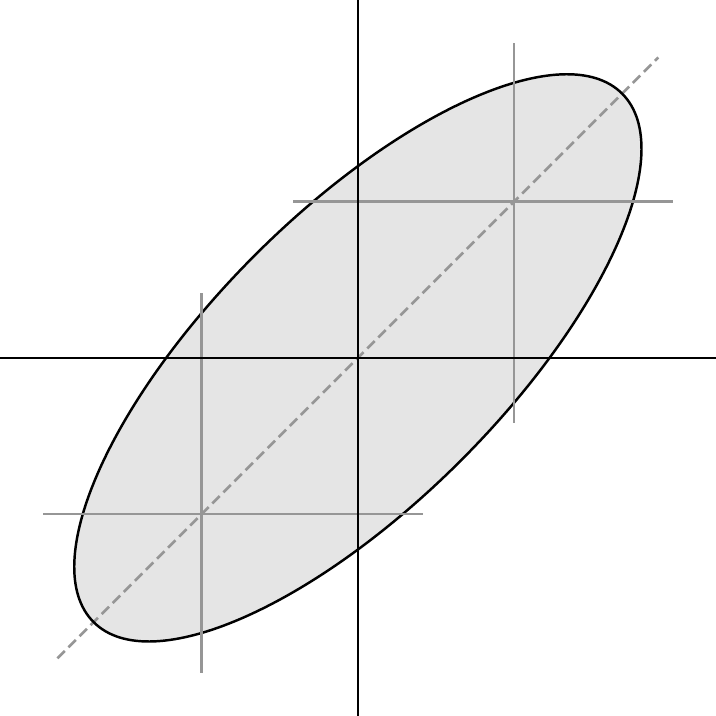} \\
\bottomrule
\end{tabular}}
\end{small}
\end{center}
  \caption{Mask criteria studied in this section, starting with \largefinal that was used in \ltp. Names we use to refer to the various methods are given along with the formula that projects each $(w_i, w_f)$ pair to a score. Weights with the largest scores (colored regions) are kept, and weights with the smallest scores (gray regions) are pruned. The $x$ axis in each small figure is $w_i$ and the $y$ axis is $w_f$. In two methods, $\alpha$ is adjusted as needed to align percentiles between $w_i$ and $w_f$. When masks are created, ties are broken randomly, so a score of 0 for every weight results in random masks.
}
\vspace*{-.5em}
\figlabel{masks}
\end{figure}

% \begin{figure}[h!]
% \begin{center}
% \includegraphics[width=1\linewidth]{pruning_methods_condensed}
% \caption{
% Test accuracy at early stopping iteration\footnote{The early stopping criterion we employ throughout this paper is the iteration of minimum validation loss during training}
% of different mask criteria for four networks at various pruning rates. 
%   Each line is a different mask criteria, with bands around \maginc, \largefinal, \movement, and \random depicting the min and max over 5 runs. Stars represent points where \largefinal or \maginc are significantly above the other at a $p < 0.05$ level.
% \largefinal and \maginc have the best performance, with \maginc having slightly higher accuracy in Conv2 and Conv4. As expected, criteria using small weight values consistently perform worse than random. See \figref{pruning_methods_combined_crop} for results on convergence speed.
% }
% \label{fig:pruning_methods_condensed}
% \end{center}
 
% \end{figure}

\figp[t]{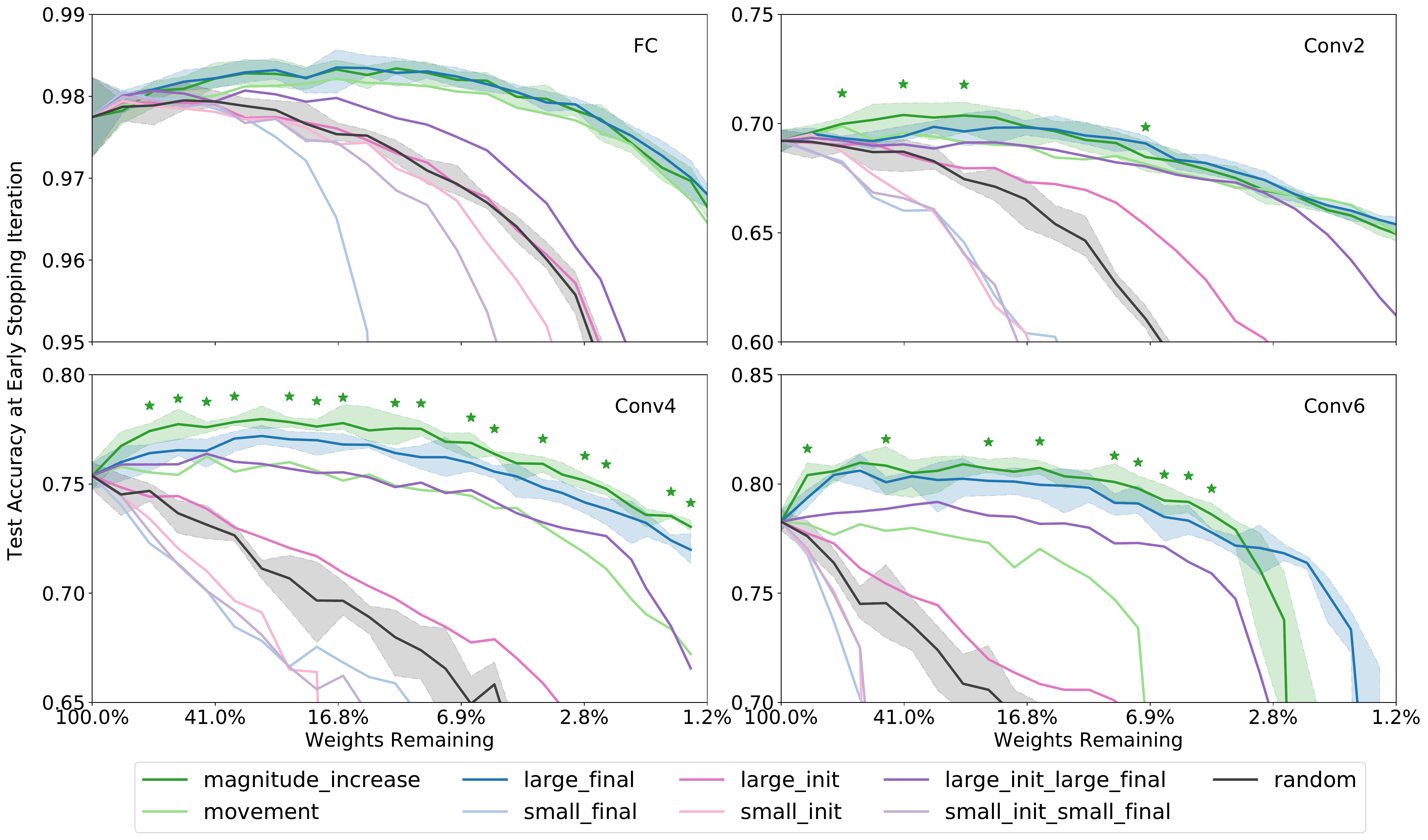}{1}{
Test accuracy at early stopping iteration
of different mask criteria for four networks at various pruning rates. Each line is a different mask criteria, with bands around the best-performing mask criteria (\largefinal and \maginc) and the baseline (\random) depicting the min and max over 5 runs. Stars represent points where \largefinal or \maginc are significantly above the other at $p < 0.05$. The eight mask criteria form four groups of inverted pairs (each column of the legend represents one such pair) that act as controls for each other. We observe that \largefinal and \maginc have the best performance, with \maginc having slightly higher accuracy in Conv2 and Conv4. 
% As expected, criteria using small weight values consistently perform worse than random. 
See \figref{pruning_methods_combined_crop} for results on convergence speed.
  \vspace*{-.5em}
  }

In this section and throughout the remainder of the paper, we follow the experimental framework from \ltp and perform iterative pruning experiments on a 3-layer fully-connected network (FC) trained on MNIST \cite{lecun-1998-IEEE-gradient-based-learning-applied} and on three convolutional neural networks (CNNs), Conv2, Conv4, and Conv6 (small CNNs with 2/4/6 convolutional layers, same as used in \ltp) trained on CIFAR-10 \cite{krizhevsky-2009-TR-learning-multiple-layers}. For more architecture and training details, see \secref{arch} in Supplementary Information. We hope to expand these experiments to larger datasets and deeper models in future work. In particular, \cite{lth_large} shows that the original LT Algorithm as proposed do not generalize to ResNet on ImageNet. It would be valuable to see how well the experiments in this paper generalize to harder problems.

Results of all criteria are shown in \figref{pruning_methods_condensed} for the four networks (FC, Conv2, Conv4, Conv6). The accuracy shown is the test accuracy at an early stopping iteration\footnote{The early stopping criterion we employ in this paper is the iteration of minimum validation loss.} of training.
For all figures in this paper, the line depicts the mean over five runs, and the band (if shown) depicts the min and max obtained over five runs. In some cases the band is omitted for visual clarity.

Note that the first six criteria as depicted in \figref{masks} form three opposing pairs; in each case, we observe when one member of the pair performs better than the random baseline, the opposing member performs worse than it. Moreover, the \maginc criterion turns out to work just as well as the \largefinal criterion, and in some cases significantly better\footnote{We run a t-test for each pruning percentage based on a sample of 5 independent runs for each mask criteria.}.

% \begin{figure}
%   \begin{center}
%   \includegraphics[width=1.0\linewidth]{pruning_methods_fc_crop}
%   \includegraphics[width=1.0\linewidth]{pruning_methods_conv2_crop}
%   \includegraphics[width=1.0\linewidth]{pruning_methods_conv4_crop}
%   \includegraphics[width=1.0\linewidth]{pruning_methods_conv6_crop}
%   \caption{
%     Mask criteria.
%     \todo{Export without dashdot lines.}
%     \todo{Add t-test stars to these figs.}
%     \todo{Refer to parking tickets}    
%     \todo{network labels?}
%   }
%   \figlabel{criteria}
%   \end{center}
% \end{figure}

The conclusion so far is that although \largefinal is a very competitive mask criterion, the LT behavior is not limited to this mask criterion as other mask criteria (\maginc, \lilf, \movement) can also match or exceed the performance of the original network. This partially answers our question about the efficacy of different mask criteria. Still unanswered: why either of the two front-running criteria (\maginc, \largefinal) should work well in the first place. We uncover those details in the following two sections.

\section{Mask-1 actions: the sign-ificance of initial weights}
\seclabel{oneaction}

Now that we have explored various ways of choosing which weights to keep and prune, we will consider how we should initialize the kept weights. In particular, we want to explore an interesting observation in \ltp which showed that the pruned, skeletal LT networks train well when you rewind to its original initialization, but degrades in performance when you randomly reinitialize the network. 

Why does reinitialization cause LT networks to train poorly? Which components of the original initialization are important? To investigate, we keep all other treatments the same as \ltp and perform a number of variants in the treatment of 1-masked, trainable weights, in terms of how to reinitialize them before the subnetwork training:
\begin{itemize}
    \item ``Reinit'' experiments: reinitialize kept weights based on the original init distribution.
    \item ``Reshuffle'' experiments: reinitialize while respecting the original distribution of remaining weights in that layer by reshuffling the kept weights' initial values.
    \item ``Constant'' experiments: reinitialize by setting 1-masked weight values to a positive or negative constant; thus every weight on a layer becomes one of three values: $-\alpha$, $0$, or $\alpha$, with $\alpha$ being the standard deviation of each layer's original initialization.
\end{itemize}

All of the reinitialization experiments are based on the same original networks and use the \largefinal mask criterion with iterative pruning. We include the original LT network (``rewind, large final'') and the randomly pruned network (``random'') as baselines for comparison.

We find that none of these three variants alone are able to train as well as the original LT network, shown as dashed lines in \figref{reinit_exp_condensed}.
However, all three variants work better when we ensure that the new values of the kept weights are of the same sign as their original initial values. These are shown as solid color lines in \figref{reinit_exp_condensed}. 
Clearly, the common factor in all working variants including the original rewind action is the sign. As long as you keep the sign, reinitialization is not a deal breaker; in fact, even setting all kept weights to a constant value consistently performs well! The significance of the sign suggests, in contrast to \cite{frankle-2019-ICLR-the-lottery-ticket-hypothesis}, that the basin of attraction for an LT network is actually quite large: optimizers work well anywhere in the correct sign quadrant for the weights, but encounter difficulty crossing the zero barrier between signs.

\figp[t]{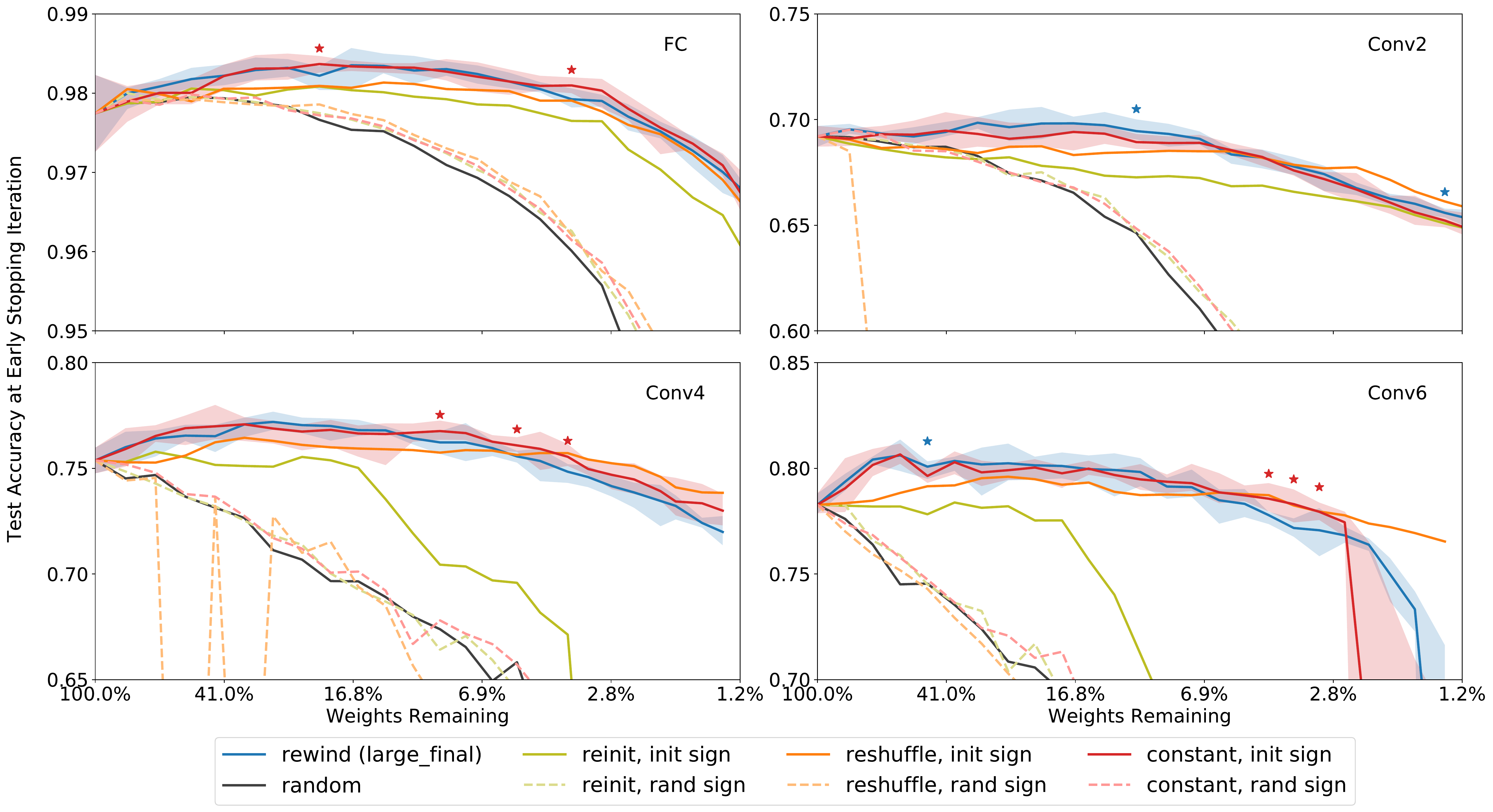}{1}{The effects of various 1-actions for four networks at various pruning rates. All reinitialization experiments use the \largefinal mask criterion with iterative pruning. Dotted lines represent the three described methods, and solid lines are those three except with each weight having the same sign as its original initialization. Shaded bands around notable runs depict the min and max over 5 runs. Stars represent points where "rewind (\largefinal)" or "constant, init sign" is significantly above the other at a $p < 0.05$ level, showing no difference in performance between the two. The original \largefinal and \random are included as baselines. See \figref{reinit_exps_combined_crop} for results on convergence speed.
\vspace*{-1.5em}
}

\section{Mask-0 actions: masking is training}
\seclabel{zeroaction}

% original version:
% What do we do with weights that are pruned?
% This question seems trivial. The simplest and most ubiquitous way, is to set them to zero. \todo{Add literature: what do people do differently?}
% The term ``pruning'' literally implies the dropping of connections by setting weights to zero. In fact, people consider those weights as unimportant. If they are truly unimportant, maybe setting their values to anything would lead to a similarly performing network. We found in this section that is not the case: zero values actually matter; a new approach to freezing works even better; and masking can be explained as a way of training.

What should we do with weights that are pruned?
This question may seem trivial, as deleting them (equivalently: setting them to zero) is the standard practice. The term ``pruning'' implies the dropping of connections by setting weights to zero, and these weights are thought of as unimportant. However, if the value of zero for the pruned weights is not important to the performance of the network, we should expect that we can set pruned weights to some other value, such as leaving them frozen at their initial values, without hurting the trainability of the network. This turns out to not be the case. We show in this section that zero values actually matter, alternative freezing approach results in better performing networks, and masking can be viewed as a way of training.

%\subsection{Why zeros matter}

% Considering the two actions taken on masked weights as done in \citet{frankle-2019-ICLR-the-lottery-ticket-hypothesis}---setting them to zero, and freezing them on subsequent trainings, it is never clear which of these two components leads to increased performance in LT networks.
% To separate the two factors, we run a simple experiment: we reproduce the LT iterative pruning experiments in which network weights are masked out in alternating train/mask/rewind cycles, but try an additional treatment: freeze masked weights at their initial values instead of at zero. If zero isn't special, both should perform similarly. We follow \citet{frankle-2019-ICLR-the-lottery-ticket-hypothesis} and train three convolutional neural networks (CNNs)---Conv2, Conv4, and Conv6, and a fully-connected (FC), on CIFAR-10. 

Typical network pruning procedures \cite{NIPS2015_5784,han-2015-ICLR-deep-compression:-compressing,li2016pruning} perform two actions on pruned weights: set them to zero, and freeze them in subsequent training (equivalent to removing those connections from the network). It is unclear which of these two components leads to the increased performance in LT networks. To separate the two factors, we run a simple experiment: we reproduce the LT iterative pruning experiments in which network weights are masked out in alternating train/mask/rewind cycles, but try an additional treatment: freeze masked weights at their initial values instead of at zero. If zero isn't special, both should perform similarly.
%We follow \ltp and train three convolutional neural networks (CNNs)---Conv2, Conv4, and Conv6, and a fully-connected (FC), on CIFAR-10. 

% Results are shown in \figref{freeze_init}, with pruning (or more correctly, ``freezing at some value'') progressing from unpruned on the left to very pruned networks on the right. 

\figref{freeze_init} shows the results for this experiment. We find that networks perform significantly better when weights are frozen specifically at zero than at random initial values. For these networks masked via the LT \largefinal criterion\footnote{\figref{kd_large_final_motion} illustrates why the \largefinal criterion biases weights that were moving toward zero during training toward zero in the mask, effectively pushing them further in the direction they were headed.}, zero would seem to be a particularly good value to set pruned weights to. At high levels of pruning, freezing at the initial values may perform better, which makes sense since having a large number of zeros means having lots of dead connections.

So why does zero work better than initial values? One hypothesis is that the mask criterion we use \emph{tends to mask to zero those weights that were headed toward zero anyway}.

To test out this hypothesis, we propose another mask-0 action halfway between freezing at zero and freezing at initialization: for any zero-masked weight, freeze it to zero if it moves toward zero over the course of training, and freeze it at its random initial value if it moves away from zero. We show two variants of this experiment in \figref{freeze_init}. In the first variant, we apply it directly as stated to zero-masked weights (to be pruned). We see that by doing so we achieve comparable performance to the original LT networks at low pruning rates and better at high pruning rates. In the second variant, we extend this action to one-masked weights too, that is, initialize every weight to zero if they move towards zero during training, regardless of the pruning action on them. 
% In other words, the difference between Variant 1 and 2 is that in Variant 2, we also change the initialization of a subset of kept weights to zero. 
We see that performance of Variant 2 is even better than Variant 1, suggesting that this new mask-0 action we found can be a beneficial mask-1 action too. These results support our hypothesis that the benefit derived from freezing values to zero comes from the fact that those values were moving toward zero anyway\footnote{Additional control variants of this experiment can be seen in Supplementary Information \secref{si:zeroaction}.}. This view on masking as training provides a new perspective on 1) why certain mask criteria work well (\largefinal and \maginc both bias towards setting pruned weights close to their final values in the previous round of training), 2) the important contribution of the value of pruned weights to the overall performance of pruned networks, and 3) the benefit of setting these select weights to zero as a better initialization for the network.

\begin{figure}[t]
  \begin{center}
  \includegraphics[width=.95\linewidth]{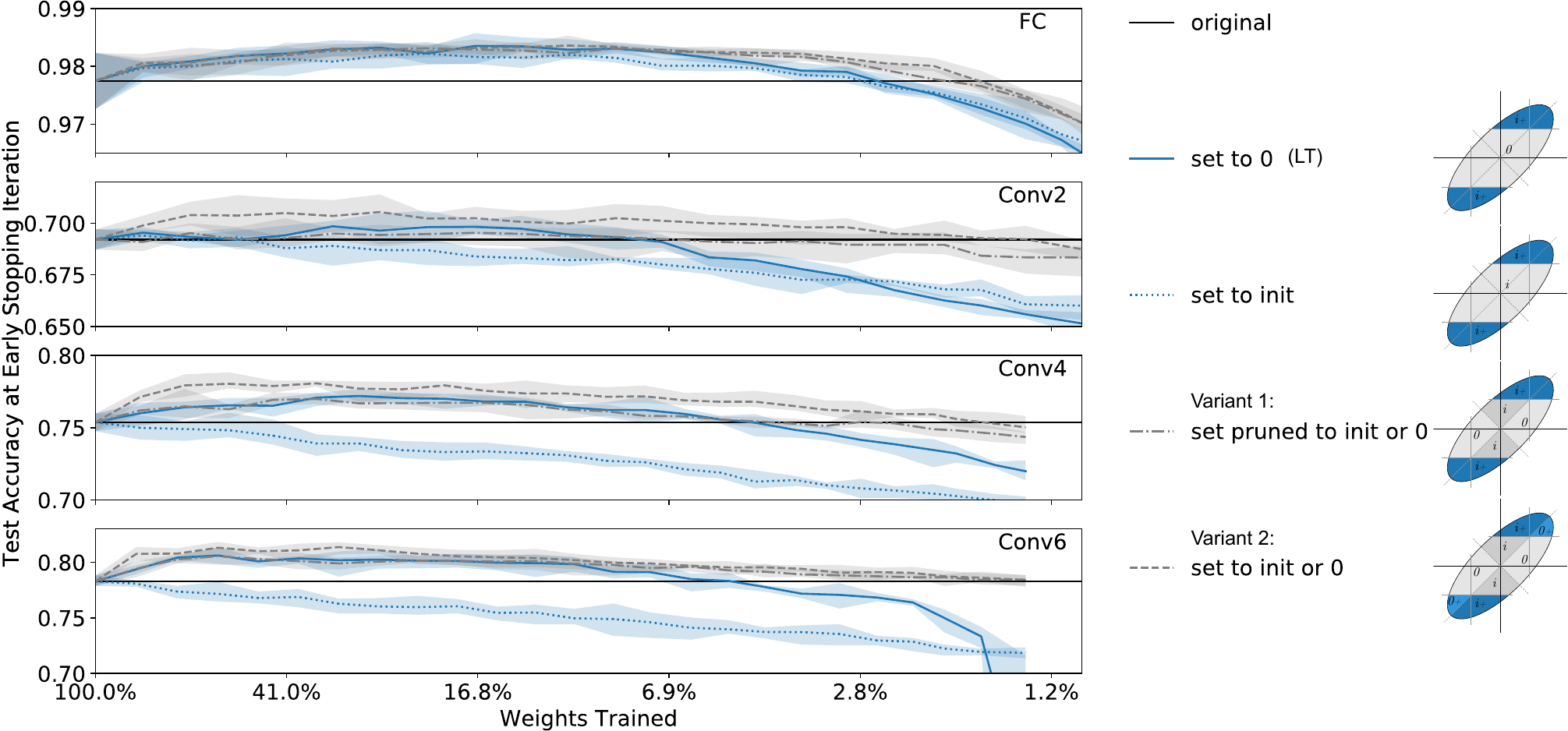}
  \caption{
     Performance of network pruning using different treatments of pruned weights (mask-0 actions). Horizontal black lines represent the performance of training the original, full network, averaged over five runs. Solid blue lines represent the original LT algorithm, which freezes pruned weights at zero. Dotted blue lines freeze pruned weights at their initial values. Grey lines show the new proposed 0-action---set to zero if they decreased in magnitude by the end of training, otherwise set to their initialization values. Two variants are shown: 1) new treatment applied to only pruned weights (dashdotted grey lines); 2) new treatment applied to all weights (dashed grey lines).
  }
  \figlabel{freeze_init}
  \end{center}
\vspace*{-1.5em}
\end{figure}

%\subsection{Masking is training}
\section{Supermasks}
\seclabel{supermask}

%\figp{better_than_chance_init_crop}{0.5}{
%    Untrained networks perform at chance (10\% accuracy, for example, on the MNIST dataset as depicted), if they are randomly initialized, or randomly initialized and randomly masked. However, applying the \largefinal mask improves the network accuracy beyond the chance level.}

The hypothesis above suggests that for certain mask criteria, like \largefinal, that masking is training: \emph{the masking operation tends to move weights in the direction they would have moved during training}. If so, just how powerful is this training operation?
To answer this, we can start from the beginning---not training the network at all, but simply applying a mask to the randomly initialized network.

It turns out that with a well-chosen mask, an untrained network can already attain a test accuracy far better than chance. This might come as a surprise, because if you use a randomly initialized and untrained network to, say, classify images of handwritten digits from the MNIST dataset, you would expect accuracy to be no better than chance (about 10\%). But now imagine you multiply the network weights by a mask containing only zeros and ones. In this instance, weights are either unchanged or deleted entirely, but the resulting network now achieves nearly 40 percent accuracy at the task! This is strange, but it is exactly what we observe with masks created using the \largefinal criterion. 

In randomly-initialized networks with \largefinal masks, it is not implausible to have better-than-chance performance since the masks are derived from the training process. The large improvement in performance is still surprising, however, since the only transmission of information from the training back to the initial network is via a zero-one mask based on a simple criterion. We call masks that can produce better-than-chance accuracy without training of the underlying weights ``Supermasks''.

% As depicted in \figref{better_than_chance_init_crop}, in randomly-initialized networks and randomly-initialized networks with random masks, neither weights nor the mask contain any information about the labels, so accuracy cannot reliably be better than chance. In randomly-initialized networks with \largefinal masks, it is not entirely implausible to have better-than-chance performance since the masks are derived from the training process. The large improvement in performance is still surprising, however, since the only transmission of information from the training back to the initial network is via a zero-one mask based on a simple criterion.
% We call these masks that can produce better-than-chance accuracy without training of the underlying weights ``Supermasks''.

We now turn our attention to finding better Supermasks.
First, we simply gather all masks instantiated in the process of creating the networks shown in \figref{masks}, apply them to the original, randomly initialized networks, and evaluate the accuracy without training the network. Next, compelled by the demonstration in \secref{oneaction} of the importance of signs and in \secref{zeroaction} of keeping large weights, we define a new \largefinalss mask criterion that selects for weights with large final magnitudes that also maintained the same sign by the end of training. This criterion, as well as the control case of \largefinalds, is depicted in \figref{better_than_chance_init_crop}. Performances of Supermasks produced by all $10$ criteria are included in \figref{supermask_combined_small_crop}, compared with two baselines: networks untrained and unmasked (\untrbase) and networks fully trained (\trbase). For simplicity, we evaluate Supermasks based on one-shot pruning rather than iterative pruning.

%\begin{wrapfigure}{r}{0.35\textwidth}
%\centering
%\includegraphics[width=1\linewidth]{better_than_chance_init_crop}
%%\vspace*{-1.5em}
%\caption{\figlabel{better_than_chance_init_crop}Untrained networks perform at chance (10\% accuracy, for example, on the MNIST dataset as depicted), if they are randomly initialized, or randomly initialized and randomly masked. However, applying the \largefinal mask improves the network accuracy beyond the chance level.}
%\end{wrapfigure}

\newcolumntype{K}{>{\centering\arraybackslash}m{0.5\linewidth}}
\begin{figure}[t]
  \vskip 0.15in
  \centering
\begin{minipage}[b]{0.4\hsize}\centering
  \includegraphics[width=1\linewidth]{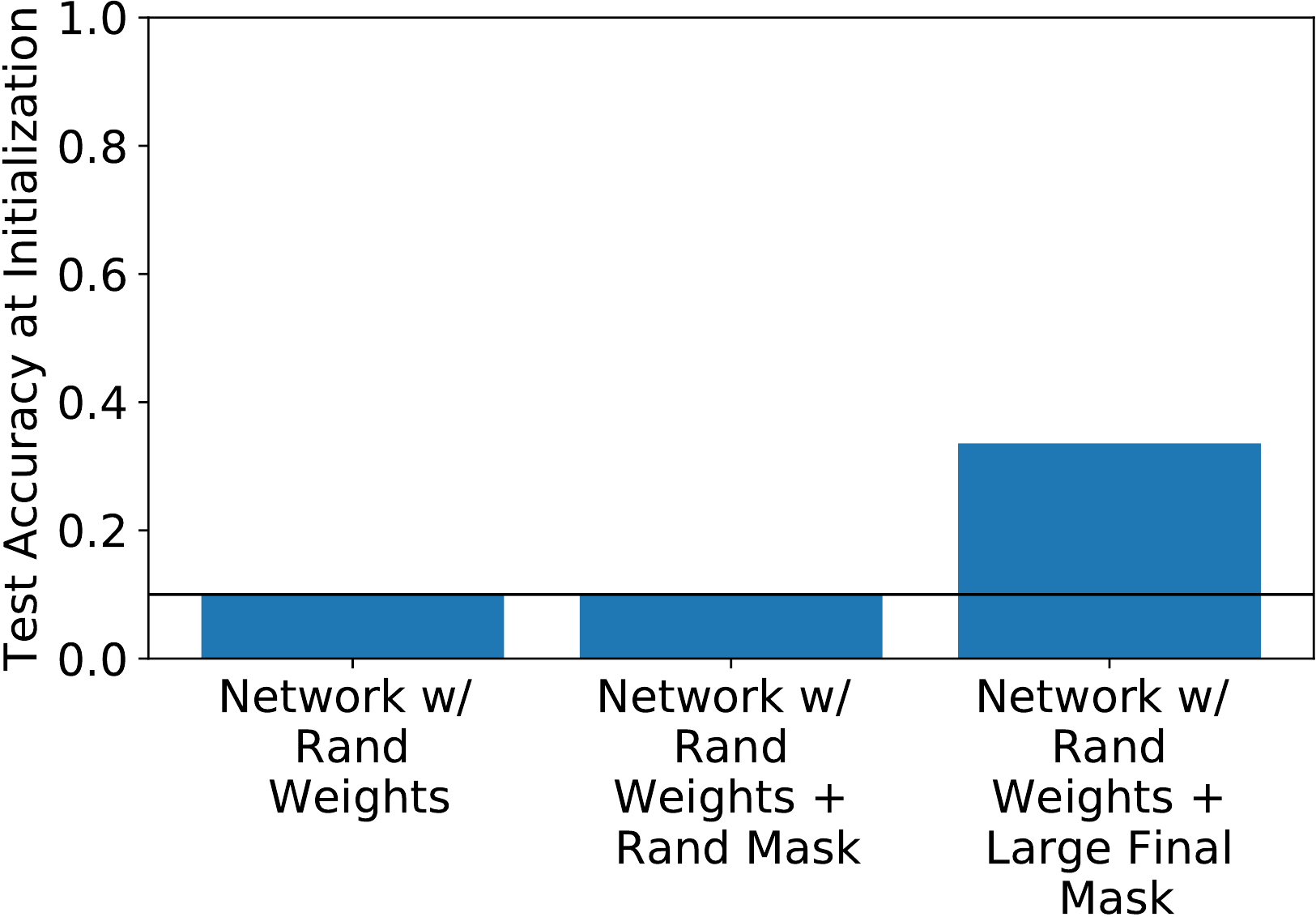}
\end{minipage}
\hspace{3em}
\begin{minipage}[b]{0.25\hsize}\centering
\begin{small}
\begin{tabular}{KK}
\toprule
large final, \vspace*{-.7em} &
large final, \vspace*{-.7em} \\
same sign &
diff sign \\
$\max(0,\frac{w_iw_f}{|w_i|})$ &
$\max(0,\frac{-w_iw_f}{|w_i|})$ \\
\includegraphics[width=1.0\linewidth]{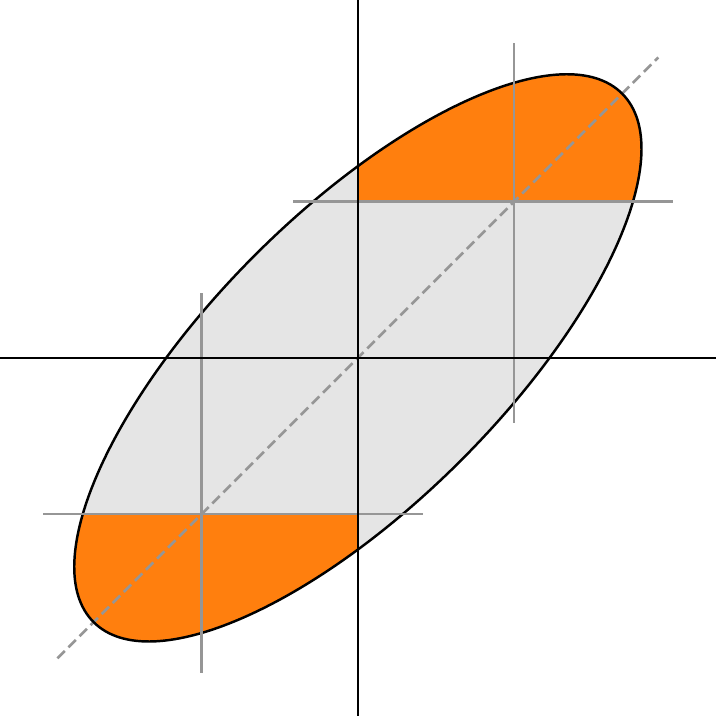} &
\includegraphics[width=1.0\linewidth]{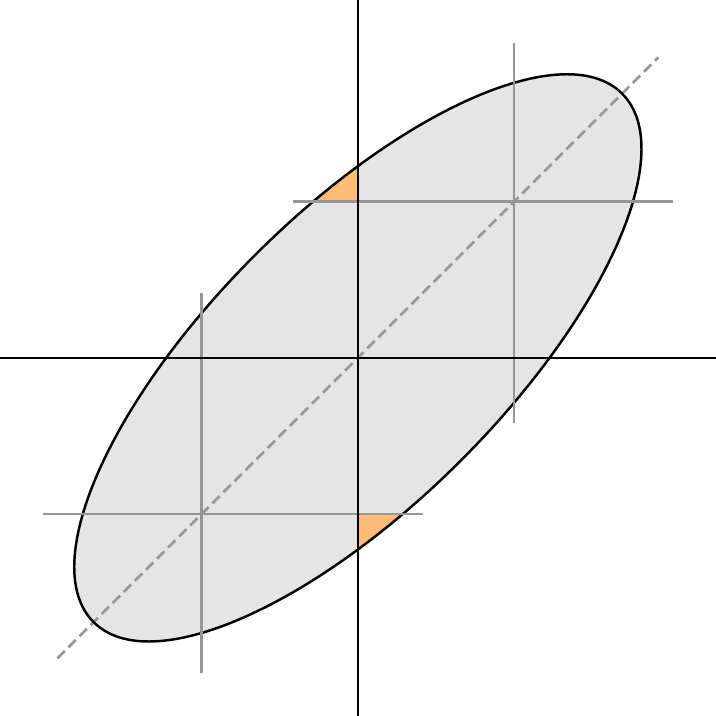} \\
\bottomrule
\end{tabular}
\end{small}
\end{minipage}
\caption{
  \capleft Untrained networks perform at chance (10\% accuracy) on MNIST, if they are randomly initialized, or randomly initialized and randomly masked. However, applying the \largefinal mask improves the network accuracy beyond the chance level.
  \capright The \largefinalss mask criterion (left) that tends to produce the best Supermasks. In contrast to the \largefinal mask in \figref{kd_large_final_detailed}, this criterion masks out the quadrants where the sign of $w_i$ and $w_f$ differ. We include \largefinalds (right) as a control.
}
\figlabel{better_than_chance_init_crop}
\vspace*{-1em}
\end{figure}

%%%%%%%%%%%%%%%%%%%%
% Figure only used for blog
%%%%%%%%%%%%%%%%%%%%
%\newcolumntype{K}{>{\centering\arraybackslash}m{0.4\linewidth}}
%\begin{table}[h]
%  \caption{The final ``large final, same sign'' mask criterion that tends to produce the best Supermasks.}
%\tablabel{better_than_chance_init_crop}
%\vskip 0.15in
%\begin{center}
%\begin{small}
%\begin{tabular}{KK}
%\toprule
%large final, same sign:~~$\max(0,\frac{w_iw_f}{|w_i|})$ \\
%\vskip .4em
%\includegraphics[width=1.0\linewidth]{kd_large_final_same_sign_detailed} \\
%\bottomrule
%\end{tabular}
%\end{small}
%\end{center}
%\end{table}

We see that \largefinalss significantly outperforms the other mask criteria in terms of accuracy at initialization. We can create networks that obtain a remarkable 80\% test accuracy on MNIST and 24\% on CIFAR-10 without training using this simple mask criterion. Another curious observation is that if we apply the mask to a signed constant (as described in \secref{oneaction}) rather than the actual initial weights, we can produce even higher test accuracy of up to 86\% on MNIST and 41\% on CIFAR-10! Detailed results across network architectures, pruning percentages, and these two treatments, are shown in \figref{supermask_combined_small_crop}. 

% The criterion \largefinalss outperforms other criteria consistently by a large margin. And the performance boost from setting mask-1 weights from original initialization value to a same-signed constant is evident, especially for CIFAR-10.

\figp{supermask_combined_small_crop}{1}{
Comparision of Supermask performances in terms of test accuracy on MNIST and CIFAR-10 classification tasks. Subfigures are across two network structures (top: FC on MNIST, bottom: Conv4 on CIFAR-10), as well as 1-action treatments (left: weights are at their original initialization, right: weights are converted to signed constants). 
No training is performed in any network. Within heuristic based Supermasks (excluding \maskcrit{learned\_mask}), the \largefinalss mask creates the highest performing Supermask by a wide margin. Note that aside from the five independent runs performed to generate uncertainty bands for this plot, every point on this plot is from the same underlying network, just with different masks. See \figref{supermask_combined_full_crop} for performance on all four networks.
\vspace*{-.5em}
}

We find it fascinating that these Supermasks exist and can be found via such simple criteria. As an aside, they also present a method for network compression, since we only need to save a binary mask and a single random seed to reconstruct the full weights of the network.

%As we were looking into why masking and re-training improves accuracy and convergence, we noticed another interesting effect of masks: by applying them, and evaluating the network immediately without any training, we can already get an impressive accuracy. For instance, with the original pruning method of final\_weight\_large, we get an accuracy of 86\% on MNIST and 41\% on CIFAR. This is quite good, considering that the only thing we've change from an init is setting things to zero.

%We can use any of the mask criteria from \figref{mask_criteria} to determine the mask, and we compare their initial accuracies in \figref{todo}. Criteria\_X does the best, which is \todo{also the best one when you retrain?}. Additionally, we found that by setting the mask-1 weights to final signed constant, it performs even better. Various mask-1 actions are compared in \figref{todo}.

%Does the initial bump explain why these pruned and retrained networks obtain better accuracy than the original network? Not quite -- though there is some correlation between how well a mask selection criteria does overall and instantly, the initial bump doesn't make a difference once the retraining has gone through enough iterations.
\vspace*{-.5em}
\subsection{Optimizing the Supermask}

We have shown that Supermasks derived using simple heuristics greatly enhance the performance of the underlying network immediately, with no training involved. In this section we are interested in how far we can push the performance of Supermasks by \emph{training the mask}, instead of training network weights. Similar works in this domain include training networks with binary weights \cite{NIPS2015_5647, courbariaux2016binarized}, or training masks to adapt a base network to multiple tasks \cite{mallya2018piggyback}. Our work differs in that the base network is randomly initialized, never trained, and masks are optimized for the original task.

% We have shown that we can derive Supermasks with remarkable performance using simple heuristics. Related work have explored training binary masks in transfer learning settings\todo{add citation for piggyback paper here: https://arxiv.org/abs/1801.06519}. In this section, we push the performance of Supermasks by learning them directly for the original task.

% One can search, in the search space of all $2^n$ possible masks, where $n$ is the number of parameters in that network.

We do so by creating a trainable mask variable for each layer while freezing all original parameters for that layer at their random initialization values.
For an original weight tensor $w$ and a mask tensor $m$ of the same shape, we have as the effective weight $w' = w_i \odot g(m)$, where $w_i$ denotes the initial values weights are frozen at, $\odot$ is element-wise multiplication and $g$ is a point-wise function that transform a matrix of continuous values into binary values.
% One example of $g$ is $\lfloor(S(m))\rceil$, where $S$ is the sigmoid function and $\lfloor \rceil$ means rounding. Bias terms are added as usual to the product of $w'$ with the inputs as per the usual fully connected or convolutional kernels.

We train the masks with $g(m) = \operatorname{Bern}(S(m))$, where $\operatorname{Bern}(p)$ is the bernoulli sampler with probability $p$, and $S(m)$ is the sigmoid function.
% It works slightly better than $\lfloor(S(m))\rceil$.
The bernoulli sampling adds some stochasticity that helps with training, mitigates the bias of all things starting at the same value, and uses in effect the expected value of $S(m)$, which is especially useful when they are close to 0.5. 

By training the $m$ matrix with SGD, we obtained up to 95.3\% test accuracy on MNIST and 65.4\% on CIFAR-10. 
Results are shown in \figref{supermask_combined_small_crop}, along with all the heuristic-based, unlearned Supermasks. Note that there is no straightforward way to control for the pruning percentage. Instead, we initialize $m$ with larger or smaller magnitudes, which nudges the network toward pruning more or less. This allows us to produce masks with the amounts of pruning (percentages of zeros) ranging from 7\% to 89\%.
Further details about the training can be seen in \secref{si:supermask}.
%\maybe{While the weights could theoretically learn to cancel that out, in practice they consistently prune to different levels, which is our desired effect. Thus we can measure performance over different prune percentages, detailed in \figref{supermask_combined_crop.pdf}.} This is similar to the optimal prune percentages from earlier experiments.

\vspace*{-.5em}
\subsection{Dynamic Weight Rescaling}

One beneficial trick in Supermask training is to dynamically rescale the values of weights based on the sparsity of the network in the current training iteration. For each training iteration and for each layer, we multiply the underlying weights by the ratio of the total number of weights in the layer over the number of ones in the corresponding mask. Dynamic rescaling leads to significant improvements in the performance of the masked networks, which is illustrated in \tabref{supermask-testacc}.

\tabref{supermask-testacc} summarizes the best test accuracy obtained through different treatments.
The result shows striking improvement of learned Supermasks over heuristic based ones. Learned Supermasks result in performance close to training the full network, which suggests that a network upon initialization already contains powerful subnetworks that work well without training.

\muchlater{Additionally, the learning of Supermask allows identifying a possibly optimal pruning rate for each layer, since each layer is free to learn the distribution of $0$s in $m$ on their own. For instance, in \ltp the last layer of each network is designed to be pruned approximately half as much as the other layers, in our setting this ratio is automatically adjusted.}

%Setting a subset of weights to zero is such a simple operation that we wondered, what is the best possible accuracy we can get with this operation? To find the optimal set of weights to set to zero, we decided to 

% To SI? 
% However, bernoulli sampling doesn't have a gradient, so we add \texttt{sigmoid(mask) - tf.stop\_gradient(sigmoid(mask))} into \texttt{f}, and that's what we train on.
% 

%\figp[t]{placeholder_supermask_adjust_prune}{0.5}{Placeholder for showing how test accuracy varies depending on how much is pruned.}
% \begin{table}[t]
%     \caption{Test accuracy of the best Supermasks with various initialization treatments. Values shown are max over any prune percentage. The left two columns show untrained network with heuristic based mask. The third shows untrained network with learned mask and dynamic scaling. The last column show performance of the fully-trained network.}
% \tablabel{supermask-testacc}
% \begin{center} \begin{small}
% %\setlength\tabcolsep{1.5pt}
% \renewcommand{\arraystretch}{1}
% \begin{tabular}{lcccc}
% \toprule
%         & mask            & mask            & learned mask &\\
%         & $\odot$         & $\odot$         & $\odot$ & \\
% Network & initial weights & signed constant & dynamic signed constant & trained network \\
% \midrule
% MNIST FC     & 79.3  & 86.3  & 98.0 & 97.7 \\
% CIFAR Conv2  & 22.3  & 37.4  & 66.0 & 69.2 \\
% CIFAR Conv4  & 23.7	 & 39.7  & 72.5 & 75.4 \\
% CIFAR Conv6  & 24.0  & 41.0  & 76.5 & 78.3 \\
% \bottomrule
% \vspace*{-1.5em}
% \end{tabular}
% \end{small} \end{center}

% \end{table}

\begin{table}[t]
  \caption{Test accuracy of the best Supermasks with various initialization treatments. Values shown are the max over any prune percentage and averaged over four or more runs. The first two columns show untrained networks with heuristic-based masks, where ``init'' stands for the initial, untrained weights, and ``S.C.'' is the signed constant approach, which replaces each random initial weight with its sign as described in \secref{oneaction}. The next two columns show results for untrained weights overlaid with learned masks; and the two after add the Dynamic Weight Rescaling (DWR) approach. The final column shows the performance of networks with weights trained directly using gradient descent. Bold numbers show the performance of the best Supermask variation.}
\tablabel{supermask-testacc}
\begin{center} \begin{small}
\renewcommand{\arraystretch}{1.0}
\begin{tabular}{lccccccc}
\toprule
        &           &           &          &          & DWR      & DWR      & \\
        &           &           & learned  & learned  & learned  & learned  & \\
        & mask      & mask      & mask     & mask     & mask     & mask     & \\
        & $\odot$   & $\odot$   & $\odot$  & $\odot$  & $\odot$  & $\odot$  & trained \\
Network & init      & S.C.      & init     & S.C.     & init     & S.C.     & weights \\
\midrule
% OLD
%MNIST FC     & 79.3  & 86.3  & 95.3 (+2.5) & 96.38 (+1.7) & 97.7 \\
%CIFAR Conv2  & 22.3  & 37.4  & 64.4 (+0.6) & 66.32 (-0.32) & 69.2 \\
%CIFAR Conv4  & 23.7	 & 39.7  & 65.4 (+6.3) & 66.24 (+6.3) & 75.4 \\
%CIFAR Conv6  & 24.0  & 41.0  & 65.3 (+11.0) & 65.38 (+11.1) & 78.3 \\
% NEW
MNIST FC     & 79.3  & 86.3  & 95.3  & 96.4  & 97.8  & \textbf{98.0}  & 97.7 \\
CIFAR Conv2  & 22.3  & 37.4  & 64.4  & \textbf{66.3}  & 65.0  & 66.0 & 69.2 \\
CIFAR Conv4  & 23.7   & 39.7  & 65.4  & 66.2  & 71.7  & \textbf{72.5}  & 75.4 \\
CIFAR Conv6  & 24.0  & 41.0  & 65.3  & 65.4  & 76.3  & \textbf{76.5} & 78.3 \\
\bottomrule
\vspace*{-1.5em}
\end{tabular}
\end{small} \end{center}
\end{table}

\section{Conclusion}

In this paper, we have studied how three components to LT-style network pruning---mask criterion, treatment of kept weights during retraining (mask-1 action), and treatment of pruned weights during retraining (mask-0 action)---come together to produce sparse and performant subnetworks. We proposed the hypothesis that networks work well when pruned weights are set close to their final values. Building on this hypothesis, we introduced alternative freezing schemes and other mask criteria that meet or exceed current approaches by respecting this basic rule. We also showed that the only element of the original initialization that is crucial to the performance of LT networks is the sign, not the relative magnitude of the weights. Finally, we demonstrated that the masking procedure can be thought of as a training operation, and
consequently we uncovered the existence of Supermasks, which can produce partially working networks without training.

\subsubsection*{Acknowledgments}

The authors would like to acknowledge
Jonathan Frankle,
Joel Lehman,
Zoubin Ghahramani,
Sam Greydanus,
Kevin Guo, 
and members of the Deep Collective research group at Uber AI
for combinations of helpful discussion, ideas, feedback on experiments, and comments on early drafts of this work.

\clearpage

%\section*{References}

\bibliography{jby_refs,other_refs}
\bibliographystyle{plain}

%%%%%%%%%%%%%%%%%%%%%%%%%%%%%%%%%%%%%%%%%%%%%%%%%%%%%%%%%%%%%%%%%%%%%
%
% SUPPLEMENTARY INFORMATION
%
%%%%%%%%%%%%%%%%%%%%%%%%%%%%%%%%%%%%%%%%%%%%%%%%%%%%%%%%%%%%%%%%%%%%%

\clearpage

\renewcommand{\thesection}{S\arabic{section}}
\renewcommand{\thesubsection}{\thesection.\arabic{subsection}}

% Issue: Resetting counters breaks the refs to the sections/figures in SI.
\newcommand{\beginsupplementary}{%
        \setcounter{table}{0}
    \renewcommand{\thetable}{S\arabic{table}}%
        \setcounter{figure}{0}
    \renewcommand{\thefigure}{S\arabic{figure}}%
        \setcounter{section}{0}
}

\beginsupplementary

\onecolumn
\noindent\makebox[\linewidth]{\rule{\linewidth}{3.5pt}}

\begin{center}
    {\LARGE \bf Supplementary Information for:\\ \titl\par}
\end{center}
\noindent\makebox[\linewidth]{\rule{\linewidth}{1pt}}

\section{Architectures and training hyperparameters}
\seclabel{arch} 

\tabref{arch} contains the architectures used in this study, together with relevant training hyperparameters, based off of experiments in \ltp. 

\paragraph{Additional details} To compare with results in \ltp, we used the same training hyperparameters and did not do any additional tuning. Training/test splits were given in both MNIST and CIFAR-10; validation was split randomly from the training set with (55000, 5000) train and val for MNIST and (45000, 5000) for CIFAR. 

Our experiments required more computation than regular training procedures, as networks were trained up to 24 times with iterative pruning. We used single GPUs for each experiment (NVIDIA GeForce GTX 1080 Ti) and parallelized by running multiple experiments on multiple GPUs.

\begin{table}[h]
  \caption{
The architectures used in this paper. Table reproduced and modified from \ltp.
Conv networks use 3x3 convolutional layers with max pooling followed by fully connected layers.
FC layer sizes are from \cite{lecun-1998-IEEE-gradient-based-learning-applied}. Initializations are
Glorot Normal \cite{glorot-2010-aistats-understanding-the-difficulty-of-training} and activations are ReLu. }
\tablabel{arch}
%\scriptsize
\centering
\begin{tabular}{@{}l@{}c@{~~~}c@{~~}c@{~~}c@{~~}c@{~~}}\toprule
\textit{Network} & MNIST FC & CIFAR-10 Conv2 & CIFAR-10 Conv4 & CIFAR-10 Conv6  \\ \midrule
\textit{Convolutional Layers} & &
\shortstack{64, 64, pool}  &
\shortstack{64, 64, pool\\128, 128, pool}  &
\shortstack{64, 64, pool\\128, 128, pool\\256, 256, pool}
\\ \midrule

\textit{FC Layers} None & 300, 100, 10 & 256, 256, 10 & 256, 256, 10 & 256, 256, 10 \\ \midrule

\textit{All/Conv Weights} & 266K & 4.3M / 38K & 2.4M / 260K & 2.3M / 1.1M  \\ \midrule

\textit{Iterations/Batch} & 50K / 60 & 20K / 60 & 25K / 60 & 30K / 60  \\ \midrule 

\textit{Optimizer} &Adam 1.2e-3 & Adam 2e-4 & Adam 3e-4 & Adam 3e-4  \\  \midrule

\textit{Pruning Rates} & fc20\% & conv10\% fc20\% & conv10\% fc20\% & conv15\% fc20\%  \\
\bottomrule
\end{tabular}
\end{table}

\section{Further mask criteria details}
\seclabel{prune_methods}

\figref{pruning_methods_combined_crop} shows the convergence speed and performance of all mask critera for FC on MNIST and Conv2, 4, 6 on CIFAR-10.

\figp[h]{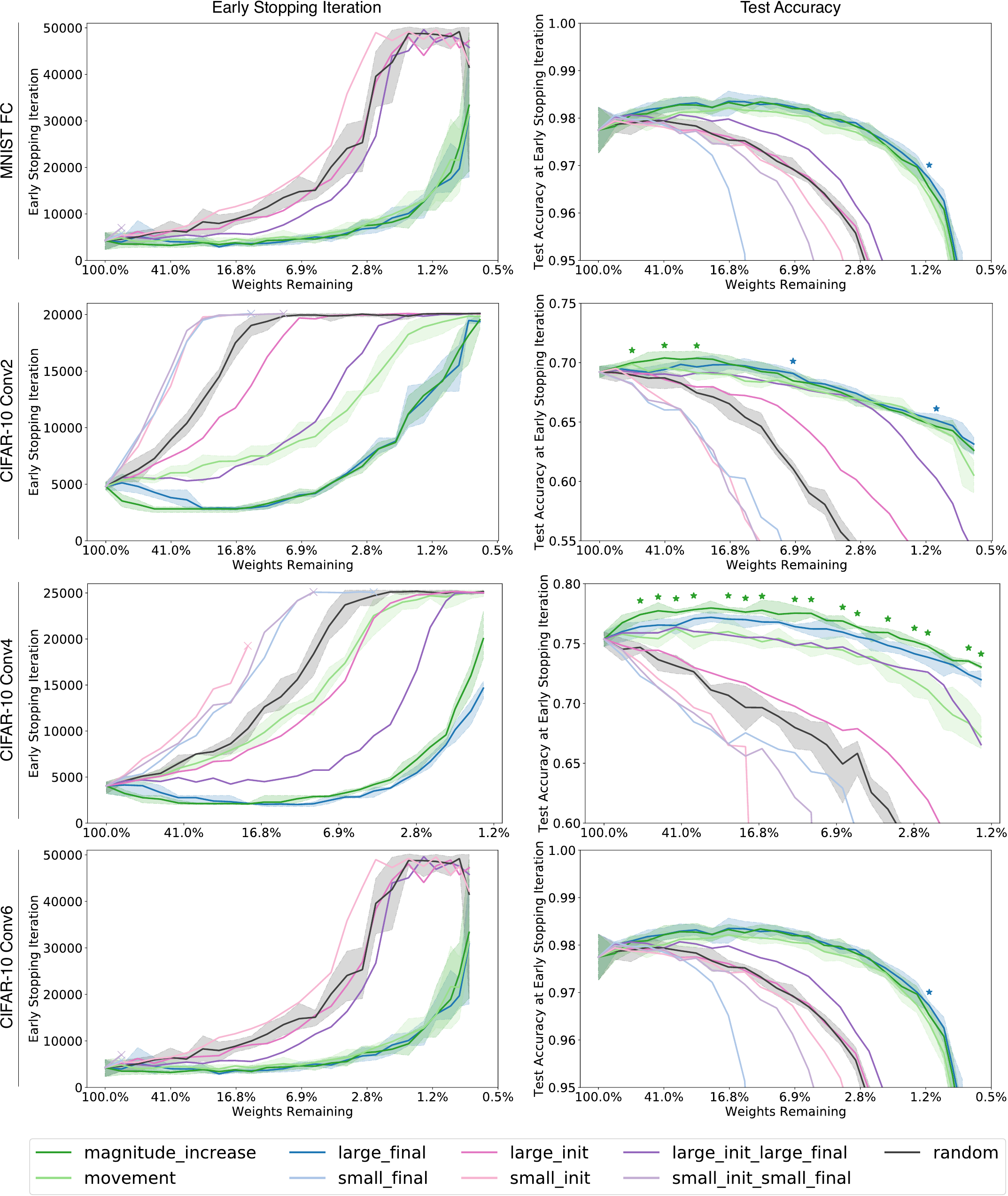}{1}{Performance of different mask criteria for four networks at various pruning rates. We show early stopping iteration on the left and test accuracy on the right.
  Each line is a different mask criteria, with bands around \maginc, \largefinal, \movement, and \random depicting the min and max over 5 runs. Stars represent points where \largefinal or \maginc are significantly above the other at a $p < 0.05$ level.
\largefinal and \maginc show the best convergence speed and accuracy, with \maginc having slightly higher accuracy in Conv2 and Conv4. As expected, criteria using small weight values consistently perform worse than random.
  \muchlater{It's hard to tell the colors apart, so use different symbols for different winners. E.g. asterisk for green, square for blue.}
}

\section{Further mask-0 action details}
\seclabel{si:zeroaction}

In this section, we discuss additional control variants of the mask-0 action experiments shown in \figref{freeze_init}. In particular, we want to know if the improvement in performance from setting only weights that move towards zero to zero is the result of this particular selection of zero weights, rather than other quirks in the treatment. We run two control cases. In the first control experiment, we randomly freeze a subset of pruned weights to zero, with the number of zero-ed weights equal to the number of zero-ed weights in "set pruned to init or 0". In the second control experiment, we set weights that \textit{moved away from zero} to zero, reversing the proposed treatment. As shown in \figref{freeze_init_extra}, both control experiments perform significantly worse than the proposed treatment ("set pruned to init or 0"), with the reversed treatment performing worse than both randomly freezing weights to zero and freezing all pruned weights at initial value. These results are consistent with our hypothesis.

\begin{figure}[t]
  \begin{center}
  \includegraphics[width=1\linewidth]{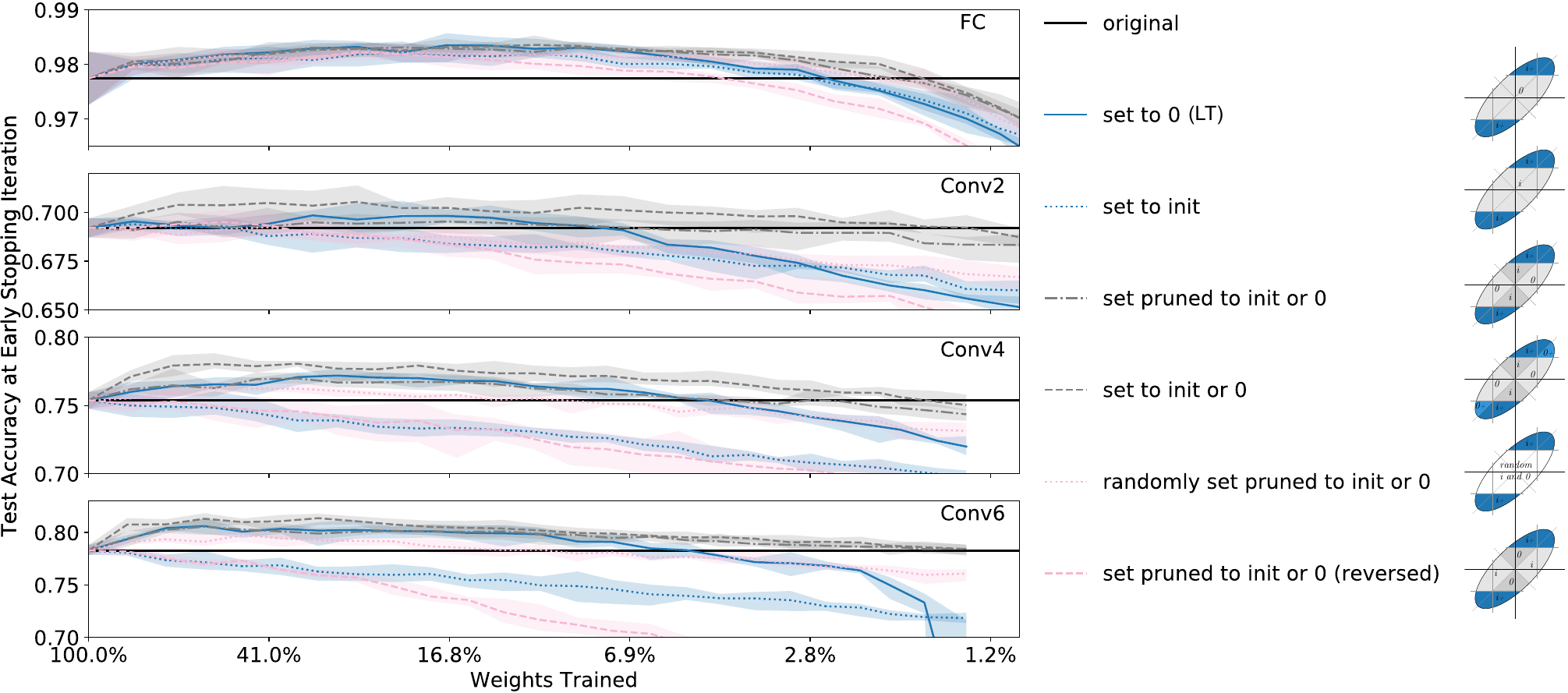}
  \caption{
     Performance of network pruning using different treatments of pruned weights (mask-0 actions). Horizontal black lines represent the performance of training the original, full network, averaged over five runs. Solid blue lines represent the original LT algorithm, which freezes pruned weights at zero. Dotted blue lines freeze pruned weights at their initial values. Grey lines show the new proposed 0-action---set to zero if they decreased in magnitude by the end of training, otherwise set to their initialization values. Two variants are shown: 1) new treatment applied to only pruned weights (dashdotted grey lines); 2) new treatment applied to all weights (dashed grey lines). Two additional control experiments are shown: 1) randomly freeze a number of pruned weights to zero, with the number equal to the number of zero-ed weights in the new proposed 0-action ("set pruned to init or 0"); 2) reverse the proposed 0-action by setting to zero weights that \textit{increased} in magnitude by the end of training.
  }
  \figlabel{freeze_init_extra}
  \end{center}
\end{figure}

%
%In the freeze\_init\_zero\_reshuffle experiment, we use standard pruning to identify pruned weights. Then, we randomly freeze the same number of pruned weights (mask-1 action in freeze\_init\_zero) to 0 and freezing the rest of the pruned weights at their initial values. This tells us that simply freezing some subset of weights to 0 and some to initial value does not explain the performance of freeze\_init\_zero experiment.
%
%
%We also try the reverse of freeze\_init\_zero whereby we freeze weights that increased in magnitude to 0 and weights that decreased to their initial values. Our hypothesis would predict that this would work far worse than the other methods. (Indeed it does!)
%
%\begin{figure}
%  \begin{center}
%  \includegraphics[width=1.0\linewidth]{freeze_init_zero_exp_fc}
%  \includegraphics[width=1.0\linewidth]{freeze_init_zero_exp_conv2}
%  \includegraphics[width=1.0\linewidth]{freeze_init_zero_exp_conv4}
%  \includegraphics[width=1.0\linewidth]{freeze_init_zero_exp_conv6}
%  \caption{
%    Freezing at 0 for shrinking weights and initial for expanding weights
%    \later{Probably change the color}
%  }
%  \figlabel{freeze_init_zero}
%  \end{center}
%\end{figure}

\figp{kd_large_final_motion}{.4}{The motion selectivity of the \largefinal mask criterion. Although \largefinal only selects for large final weights, not directly for weight motion, it biases towards pruning weights that \emph{decreased} in magnitude, since those weights are more likely to have small final magnitudes. As shown by the two paired sets of example weights, those two weights that increased during training (two solid up arrows) are reset to their initial value (dotted arrows) after pruning, resulting in no net motion. In contrast, those two weights that decrease in value during training (two solid down arrows) are set to zero (dotted lines) during pruning, moving them preferentially in the direction they moved during training. This explains why Supermasks may be created by such a simple criterion as \largefinal and motivates the more specifically designed, higher performing \largefinalss criterion (depicted in \figref{better_than_chance_init_crop}).
}

%\section{LT mask as initialization}
%
%Perhaps the most surprising property of the lottery tickets is that they improve the performance of the network. Is it simply a better initialization (from setting certain weights to 0), or does the freezing of those weights contribute to the improvement? We look at two experiments where we initialize the network the same way as a pruned network would start out, but train the entire network as before. We see that this method captures almost all of the improvement in performance seen in the lottery ticket networks.
%
%\begin{figure}
%  \begin{center}
%  \includegraphics[width=1.0\linewidth]{freeze_init_zero_original_exp_fc}
%  \includegraphics[width=1.0\linewidth]{freeze_init_zero_original_exp_conv2}
%  \includegraphics[width=1.0\linewidth]{freeze_init_zero_original_exp_conv4}
%  \includegraphics[width=1.0\linewidth]{freeze_init_zero_original_exp_conv6}
%  \caption{
%    Better initialization.
%    \later{Probably change the color}
%  }
%  \figlabel{freeze_init_zero_original}
%  \end{center}
%\end{figure}

\section{Further mask-1 action details}
\seclabel{si:oneaction}

\figref{reinit_exps_combined_crop} shows the convergence speed and performance of various reinitialization methods for FC on MNIST and Conv2, 4, 6 on CIFAR-10.

\figp[t]{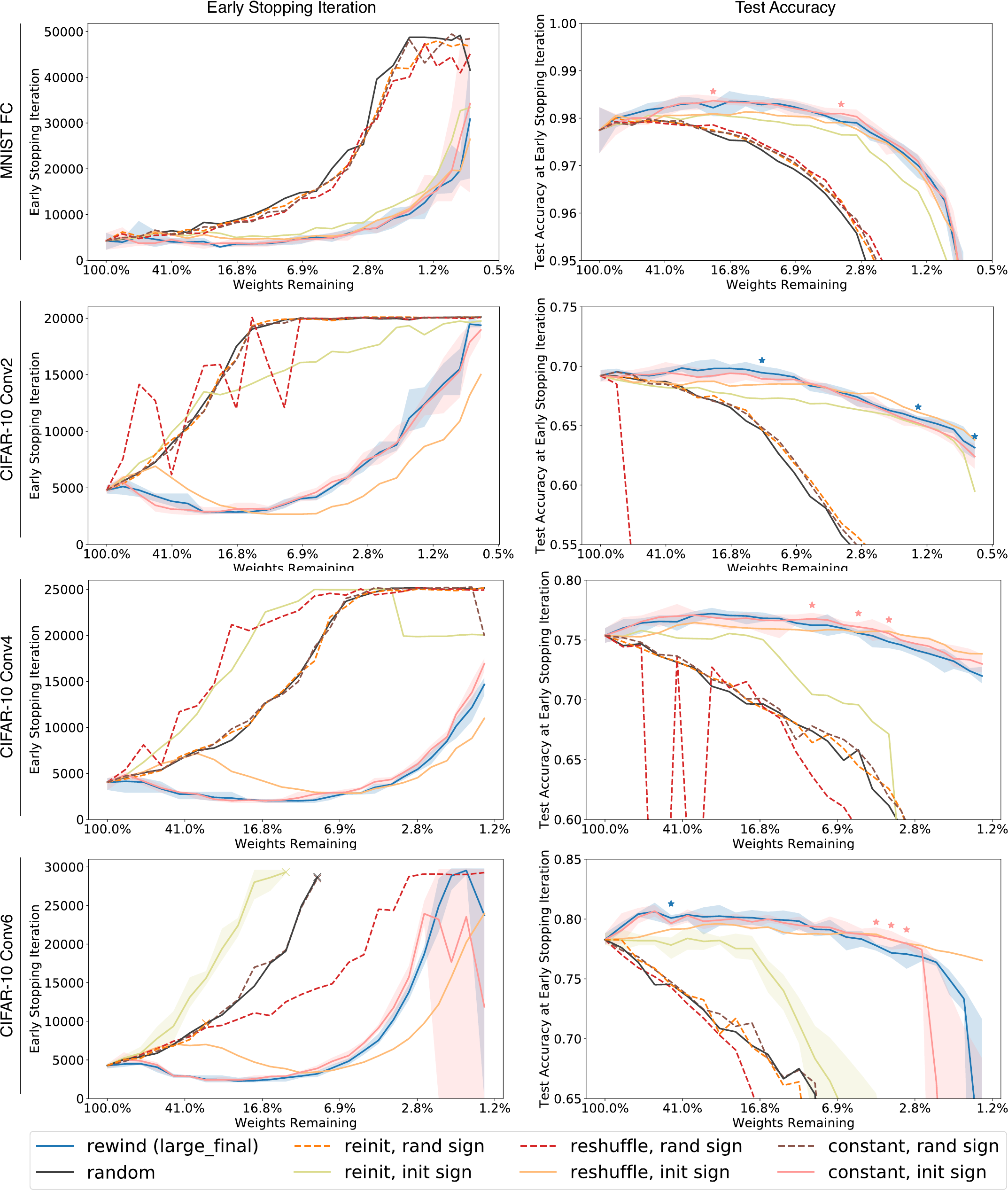}{1}{The effects of various 1-actions for the four networks and various pruning rates. Dotted lines represent the three described methods, and solid lines are those three except with each weight having the same sign as its original initialization. Shaded bands around notable runs depict the min and max over 5 runs. Stars represent points where "rewind (\largefinal)" or "constant, init sign" is significantly above the other at a $p < 0.05$ level, showing no difference in performance between the two. We also include the original rewinding method and \random pruning as baselines. ``Reshuffle, init sign'' and ``constant, init sign'' perform similarly to the ``rewind'' baseline.}

\section{Further heuristic Supermask details}
\seclabel{si:heur_masks}

We show the performance of trained LT networks with mask criteria \largefinalss and \largefinalds in \figref{large_final_sign_training_all_networks}. We see that despite having better test accuracy at iteration 0, \largefinalss slightly underperforms \largefinal when the underlying network is fully trained. This may be due to the fact that the \largefinalss mask criterion randomly select weights that changed signs to keep when there are not enough same sign weights.

\figp{large_final_sign_training_all_networks}{1}{
Test accuracy at early stopping iteration
of additional mask criteria (\largefinal, \largefinalss, \largefinalds) for four networks at various pruning rates. 
}

\section{Further training details for learning Supermasks}
\seclabel{si:supermask}

We train the networks with mask $m$ for each layer (and all regular kernels and biases frozen) with SGD, 0.9 momentum. The \{FC, Conv2, Conv4, Conv6\} networks respectively had \{100, 100, 50, 20\} for learning rates and trained for \{2000, 2000, 1000, 800\} iterations. These hyperparameters may seem absurd, but a network of masks is quite different and cannot train well with typical learning rates. Conv4 and Conv6 showed significant overfitting, thus we used early stopping as we are unable to use standard regularizing techniques. For evaluation, we also use Bernoulli sampling, but average the accuracies over 10 independent samples.

For adjusting the amount pruned, we initialized $m$ in every layer to be the same constant, which ranged from -5 to 5. In the future it may be worth trying different initializations of $m$ for each layer for more granular control over per-layer pruning rates. A different method to try would be to add an L1 loss to influence layers to go toward certain values, which may alleviate the cold start problems of some networks not learning anything due to mask values starting too low (effectively having the entire network start at zero).

\figp{supermask_combined_full_crop}{1}{
Comparision of Supermask performances in terms of test accuracy on MNIST and CIFAR-10 classification tasks. Subfigures are across various network structures (from top row to bottom: FC on MNIST, Conv2 on CIFAR-10, Conv4 on CIFAR-10, Conv6 on CIFAR-10), as well as 1-action treatments (left: weights are frozen at their original initialization, right: weights are frozen at a signed constant). 
No training is performed in any network. Weights are frozen at either initialization or constant and various masks are applied. Within heuristic based Supermasks (excluding \maskcrit{learned\_mask}), the \largefinalss mask creates the highest performing Supermask by a wide margin. Note that aside from the five independent runs performed to generate uncertainty bands for this plot, every data point on the plot is the same underlying network, just with different masks.
}

%% Extra fig for blog
%%
%\newcolumntype{J}{>{\centering\arraybackslash}m{0.095\linewidth}}
%\begin{figure}[t]
%  \vskip 0.15in
%\begin{center}
%\begin{small}
%{\setlength{\tabcolsep}{4pt}
%\hspace*{-.75em}\begin{tabular}{JJJJJJJJJJJ}
%\toprule
%large final &
%small final &
%large init &
%small init &
%large init large final &
%small init small final &
%magnitude increase &
%movement &
%large final, same sign &
%large final, diff sign \\
%$|w_f|$ &
%$-|w_f|$ &
%$|w_i|$ &
%$-|w_i|$ &
%\scalebox{.65}{$min(\alpha|w_f|, |w_i|)$} &
%\scalebox{.60}{$-max(\alpha|w_f|, |w_i|)$} &
%\scalebox{.9}{$|w_f|-|w_i|$} &
%$|w_f - w_i|$ &
%\scalebox{.7}{$\max(0,\frac{w_iw_f}{|w_i|})$} &
%\scalebox{.7}{$\max(0,\frac{-w_iw_f}{|w_i|})$} \\
%%
%\includegraphics[width=1.0\linewidth]{kd_large_final} &
%\includegraphics[width=1.0\linewidth]{kd_small_final} &
%\includegraphics[width=1.0\linewidth]{kd_large_init} &
%\includegraphics[width=1.0\linewidth]{kd_small_init} &
%\includegraphics[width=1.0\linewidth]{kd_large_init_large_final} &
%\includegraphics[width=1.0\linewidth]{kd_small_init_small_final} &
%\includegraphics[width=1.0\linewidth]{kd_magnitude_increase} &
%\includegraphics[width=1.0\linewidth]{kd_movement} &
%\includegraphics[width=1.0\linewidth]{kd_large_final_same_sign} &
%\includegraphics[width=1.0\linewidth]{kd_large_final_diff_sign} \\
%\bottomrule
%\end{tabular}}
%\end{small}
%\end{center}
%  \caption{Figure for blog.}
%\figlabel{blogheader}
%\end{figure}

\end{document}

%% file: authors.tex
\author{
  Hattie Zhou      \vspace*{-.15em}\\
  Uber             \vspace*{-.3em}\\
  \texttt{hattie@uber.com}
  \And
  Janice Lan       \vspace*{-.15em}\\
  Uber AI          \vspace*{-.3em}\\
  \texttt{janlan@uber.com}
  \And
  Rosanne Liu      \vspace*{-.15em}\\
  Uber AI          \vspace*{-.3em}\\
  \texttt{rosanne@uber.com}
  \And
  Jason Yosinski   \vspace*{-.15em}\\
  Uber AI          \vspace*{-.3em}\\
  \texttt{yosinski@uber.com}
}